\documentclass[runningheads]{llncs}
\usepackage{graphicx}
\usepackage{amsmath,amssymb,amsfonts}
\usepackage{enumerate}
\usepackage{subfigmat}

\begin{document}
\title{Particle Swarms Reformulated towards\\a Unified and Flexible Framework}
\titlerunning{Particle Swarms Reformulated}

\author{Mauro Sebasti\'an Innocente\orcidID{0000-0001-8836-2839}}
\authorrunning{M.S. Innocente}

\institute{Autonomous Vehicles \& Artificial Intelligence Laboratory (AVAILab),\\
	Coventry University, Coventry, UK\\
\email{Mauro.S.Innocente@coventry.ac.uk}\\
\url{https://www.availab.org/}}

\maketitle              
%
\begin{abstract}

	The Particle Swarm Optimisation (PSO) algorithm has undergone countless modifications and adaptations since its original formulation in 1995. Some of these have become mainstream whereas many others have not been adopted and faded away. Thus, a myriad of alternative formulations have been proposed to the extent that the question arises as to what the basic features of an algorithm must be to belong in the PSO family. The aim of this paper is to establish what defines a PSO algorithm and to attempt to formulate it in such a way that it encompasses many existing variants. Therefore, different versions of the method may be posed as settings within the proposed unified framework. In addition, the proposed formulation generalises, decouples and incorporates features to the method providing more flexibility to the behaviour of each particle. The closed forms of the trajectory difference equation are obtained, different types of behaviour are identified, stochasticity is decoupled, and traditionally global features such as sociometries and constraint-handling are re-defined as particle's attributes.


\end{abstract}

\section{Introduction}
\label{sec:1.Intro}

Inspired by earlier bird flock simulations, the Particle Swarm Optimisation (PSO) method was proposed in 1995 \cite{KenEbe95}. It is a global optimiser in the sense that it is able to escape poor suboptimal attractors by means of a swarm of particles performing a parallel collaborative search. In fact, it is a search rather than optimisation method, as no optimality criteria is checked or guaranteed.

The overall system behaviour emerges from a combination of each particle's individual and social behaviours \cite{SieInn10}. The former is manifested by the trajectory of a particle pulled by its attractors, governed by a second order difference equation with three control coefficients. In the classical (and in most) versions of the algorithm, there is one individual attractor given by the particle's best experience, and one social attractor given by the best experience in its neighbourhood. The social behaviour is governed by the way the individually acquired information is shared among particles and therefore propagated throughout the swarm, which is controlled by the neighbourhood topology. The individual and social behaviours interact through the update of the social attractor. Thus, the two main features of the algorithm are the \textit{trajectory difference equation} (and the setting of its coefficients) and the \textit{neighbourhood topology} (a.k.a. \textit{sociometry}).

In the early days, numerous empirical studies were carried out to investigate the influence of the coefficients in the \textit{trajectory difference equation} on the overall performance of the method, and to provide guidelines for their settings \cite{ShiEbe98b,ShiEbe99,EbeShi00,KwoETAL06}. Early theoretical work \cite{OzcMoh98,OzcMoh99,Ber02,CleKen02,Tre03} provided insight into the inner workings of the method and some interesting findings of practical use such as Clerc et al.'s constriction factor(s) \cite{CleKen02} to ensure convergence. These pioneering studies were a source of inspiration and set the foundations for an explosion of theoretical work. \cite{ZheETAL03a,Ken05,Cle06b,KadSelFle06,JiaLuoYan07a,PolBro07,LiuLiuShe07,PolETAL07,BlaBra08,Cle08,Pol09,FerGar09,SpeGreSpe10,Cle10,Inn10,InnSie10,CamFasPin10,FerGar11,CleEng14,GarFer14,CleEng15,BonMic17b}).
For recent reviews of the PSO method, refer to \cite{BonMic17a,FreETAL20}.

\subsection{Trajectory Difference Equation}
\label{sec:2.DiffEq}

In \textit{classical PSO} (CPSO), three forces govern a particle's trajectory: the inertia from its previous displacement, the attraction to its own best experience, and the attraction to the best experience in its neighbourhood. The importance awarded to each of them is controlled by three coefficients: the inertia ($\omega$), the individuality ($iw$), and the sociality ($sw$) weights. Stochasticity is introduced to enhance exploration via random weights applied to $iw$ and $sw$. The behaviour of a particle, and by extension of the PSO algorithm as a whole, is very sensitive to the settings of these control coefficients. The system of two $1^\text{st}$-order difference equations for position and velocity updates in the CPSO algorithm proposed in \cite{ShiEbe98a} is rearranged in \eqref{eq:CPSO} as a single $2^\text{nd}$-order \textit{Trajectory Difference Equation}:

\begin{equation}
	\begin{split}
		\label{eq:CPSO}
		x_{ij}^{(t+1)} = x_{ij}^{(t)} &+ \omega_{ij}^{(t)} \left(x_{ij}^{(t)} - x_{ij}^{(t-1)}\right)\\
		&+ iw_{ij}^{(t)} U_{(0,1)} \left(xb_{ij}^{(t)}-x_{ij}^{(t)}\right) + sw_{ij}^{(t)} U_{(0,1)} \left(xb_{kj}^{(t)}-x_{ij}^{(t)}\right)
	\end{split}
\end{equation}
where $x_{ij}^{(t)}$ is the coordinate $j$ of the \textit{position} of particle $i$ at time-step $t$; $xb_{ij}^{(t)}$ is the coordinate $j$ of the \textit{best experience} of particle $i$ by time-step $t$; $k$ is the index identifying the particle with the best experience in the neighbourhood of particle $i$ at time-step $t$; $\omega$, $iw$ and $sw$ are the inertia, individuality, and sociality weights, respectively (which may depend on $i$, $j$, $t$); and $U_{(0,1)}$ is a random number from a uniform distribution within [0,1] resampled anew every time it is referenced.

In the original formulation \cite{KenEbe95}, $\omega = 1$ and $iw = sw = 2$. This leads to an unstable system, as particles tend to diverge. The first strategy to prevent this was to bound the size of each component of a particle's displacement, which helps prevent the so-called \textit{explosion} but does not ensure convergence or a fine-grain search. Instead, the coefficients in \eqref{eq:CPSO} can be set to ensure that.

Alternatively, Clerc et al. \cite{CleKen02} analysed the trajectory of a deterministic particle in the original PSO ($\omega = 1$) and developed so-called constriction factors ($\chi$) that ensure convergence. Some authors include both $\omega$ and $\chi$ in their formulations \cite{InnSie11,TriGho16}.

\subsection{Neighbourhood Topology}
\label{sec:2.Neighbourhood}

\begin{figure}[h!]
	\begin{subfigmatrix}{3} 
		\subfigure[Global]{\includegraphics[width=0.26\textwidth]{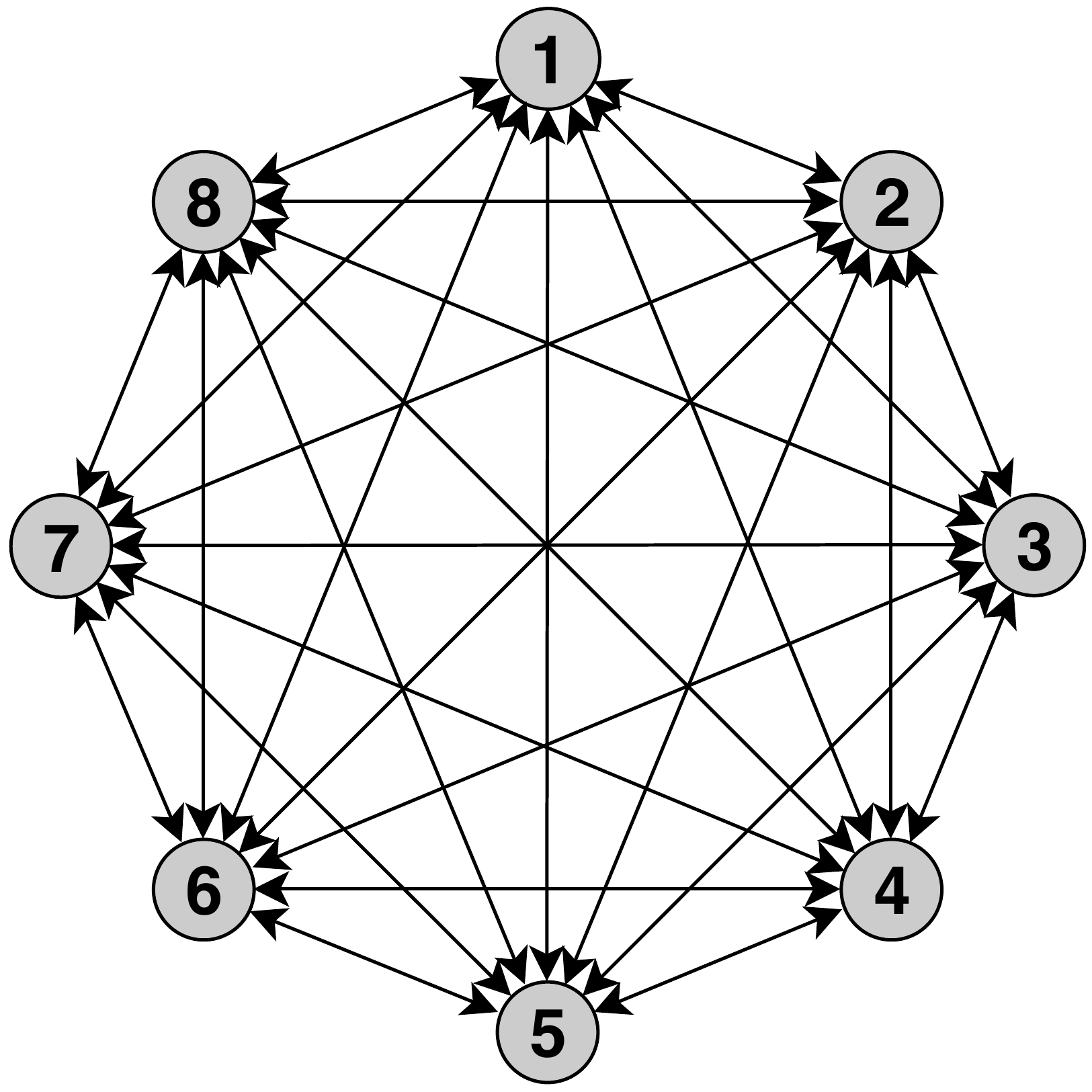}}
		\subfigure[Ring]{\includegraphics[width=0.26\textwidth]{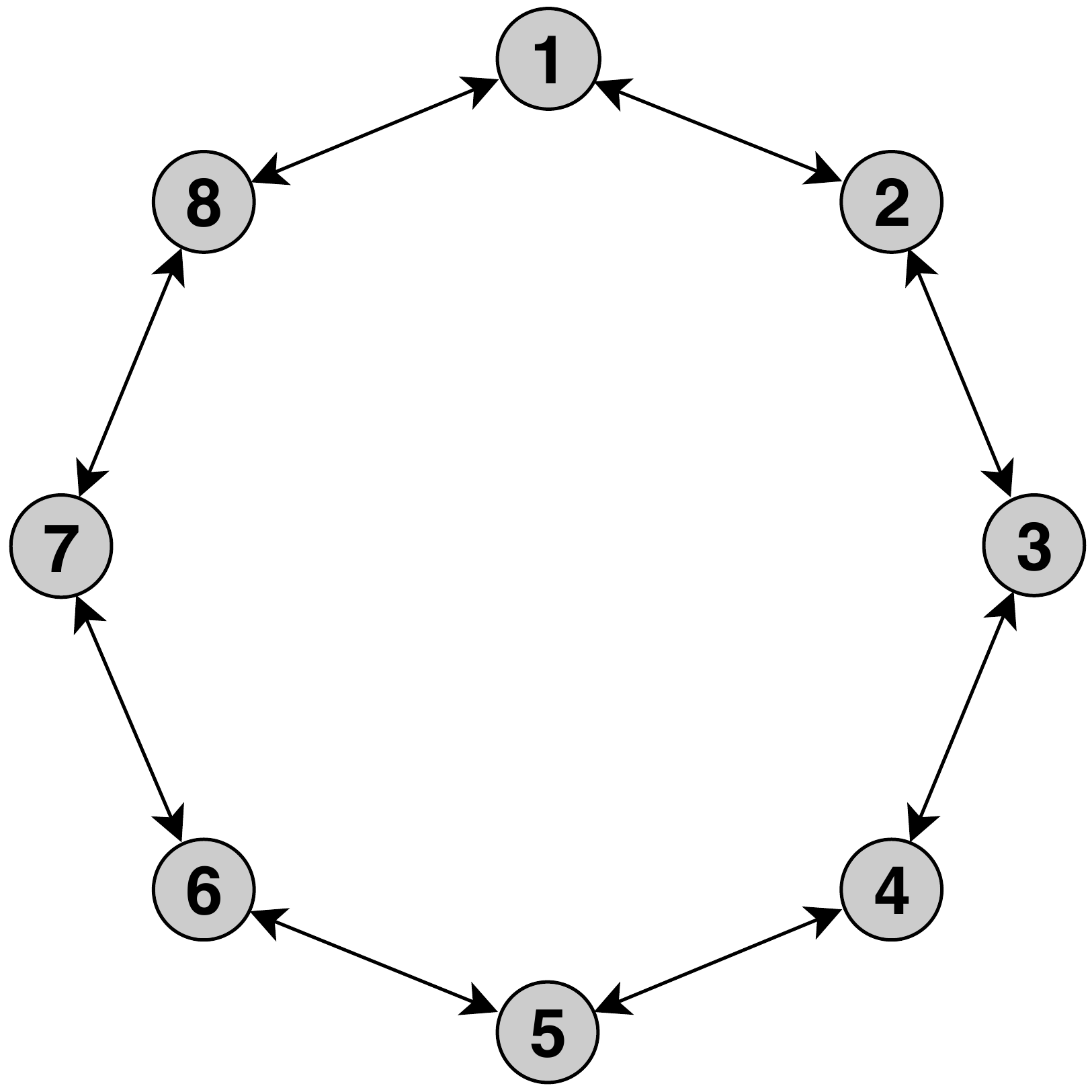}}
		\subfigure[Wheel]{\includegraphics[width=0.26\textwidth]{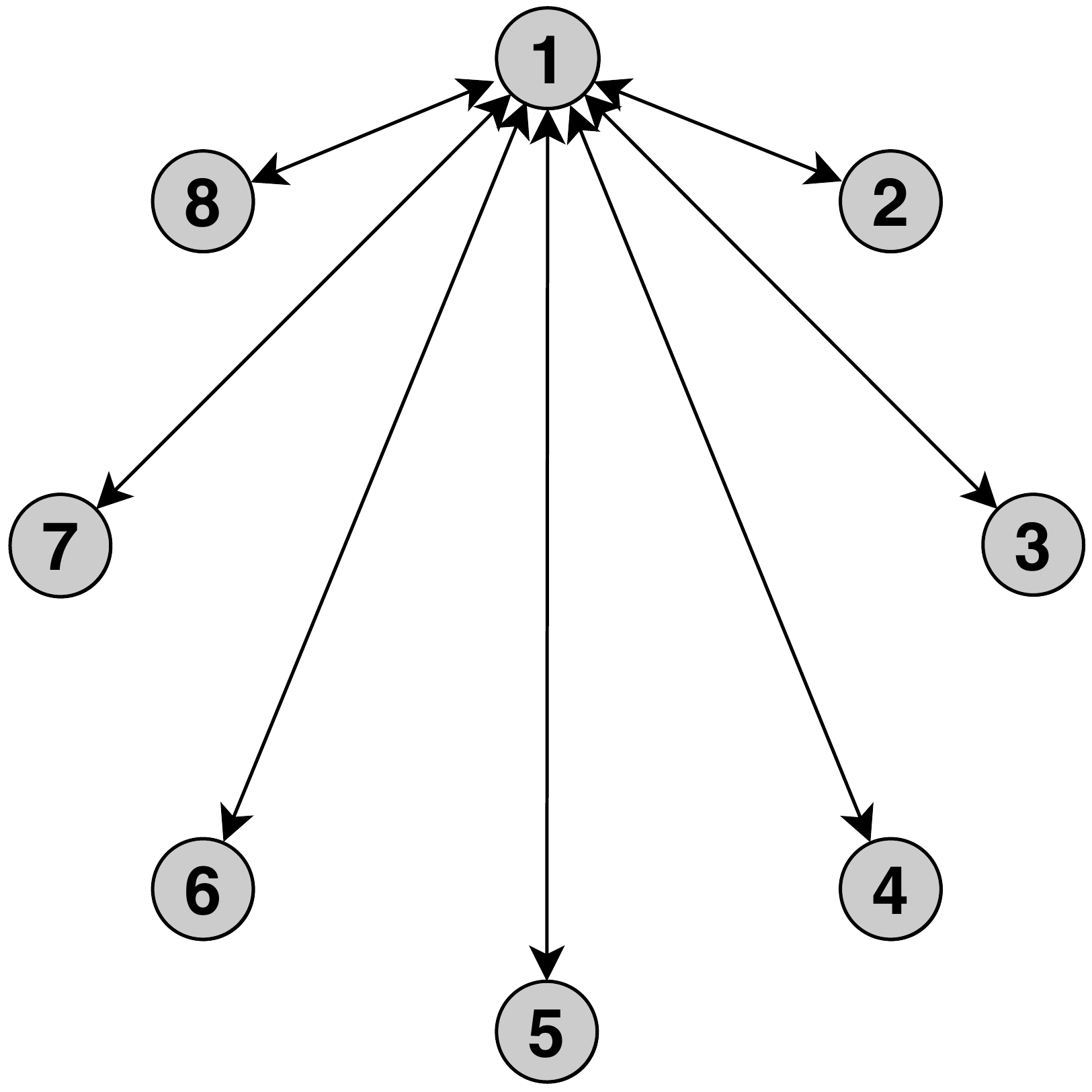}}
	\end{subfigmatrix}
	\caption{Three classical neighbourhood topologies in PSO.}
	\label{fig:Topologies}
\end{figure}

The original PSO algorithm \cite{KenEbe95} presented a global topology in which every particle has access to the  memory of every other particle in the swarm. Local topologies were proposed soon thereafter \cite{EbeKen95}. Since then, a plethora of sociometries have been proposed \cite{Men04,LiuETAL16,LynETAL18,BlaKen19}. Three classical ones are shown in Fig.~\ref{fig:Topologies}.

The global topology tends to lead to a rapid loss of diversity, which may lead to premature convergence to a poor suboptimal solution. Whilst this can be controlled to some extent by the settings of the coefficients in the trajectory equation, numerous neighbourhood topologies have been proposed reducing connectivity to delay the propagation of information throughout the swarm.

\subsection{Other Features}
\label{2:OtherFeatures}

Other important features of the PSO algorithm are the initialisation of the particles \cite{Cle08,HelWan08,KazETAL14,KazETAL14b}, the synchrony of the memory updates \cite{Azi14}, the size of the swarm \cite{DhaETAL19,PioETAL20}, and the handling of constraints \cite{Inn10,Jor15}.

The PSO algorithm is an unconstrained search method, therefore requiring an external constraint-handling technique (CHT) to be integrated to handle these types of problems. A straightforward CHT is the \textit{Preserving Feasibility Method} \cite{HuEbe02}, in which infeasible experiences are banned from memory. Another one is the \textit{Penalty Method}, in which infeasible solutions are penalised by augmenting the objective function and treating the problem as unconstrained. Some authors propose adaptive penalties by using adaptive coefficients in the penalty function \cite{Coe00} or by adapting the tolerance relaxation \cite{InnSie10b}. Innocente et al. \cite{InnETAL15} propose using a \textit{Preserving Feasibility with Priority Rules Method}, in which the objective function values and the constraint violations are treated separately.
\\
\\
Since its original formulation in 1995, countless variants have been proposed. Some of them have become mainstream whereas many others have faded away. Thus, a myriad of alternative formulations have been proposed raising the question of what the basic features of an algorithm must be to belong in the PSO family. The aim of this paper is to establish what defines a PSO algorithm, and to attempt to formulate it in such a way that it encompasses many existing variants so that different versions may be posed as settings within the proposed unified framework. In addition, the proposed formulation generalises, decouples and incorporates new features providing more flexibility to the behaviour of each particle. The remainder of this paper is organised as follows: the overall proposed \textit{Reformulated PSO} is introduced in Section~\ref{sec:1.RePSO}, with the \textit{Global Features}, the \textit{Individual Behaviour Features} and the \textit{Social Behaviour Features} discussed in more details in Sections~\ref{sec:1.GLobFeat}, \ref{sec:1.IBFeat} and \ref{sec:1.SBFeat}, respectively. Conclusions are provided in Section~\ref{sec:1.Conclusions}.

\section{Reformulated Particle Swarm Optimisation}
\label{sec:1.RePSO}

\begin{figure*}[ht!]
	\begin{center}
		\includegraphics[width=0.9\textwidth]{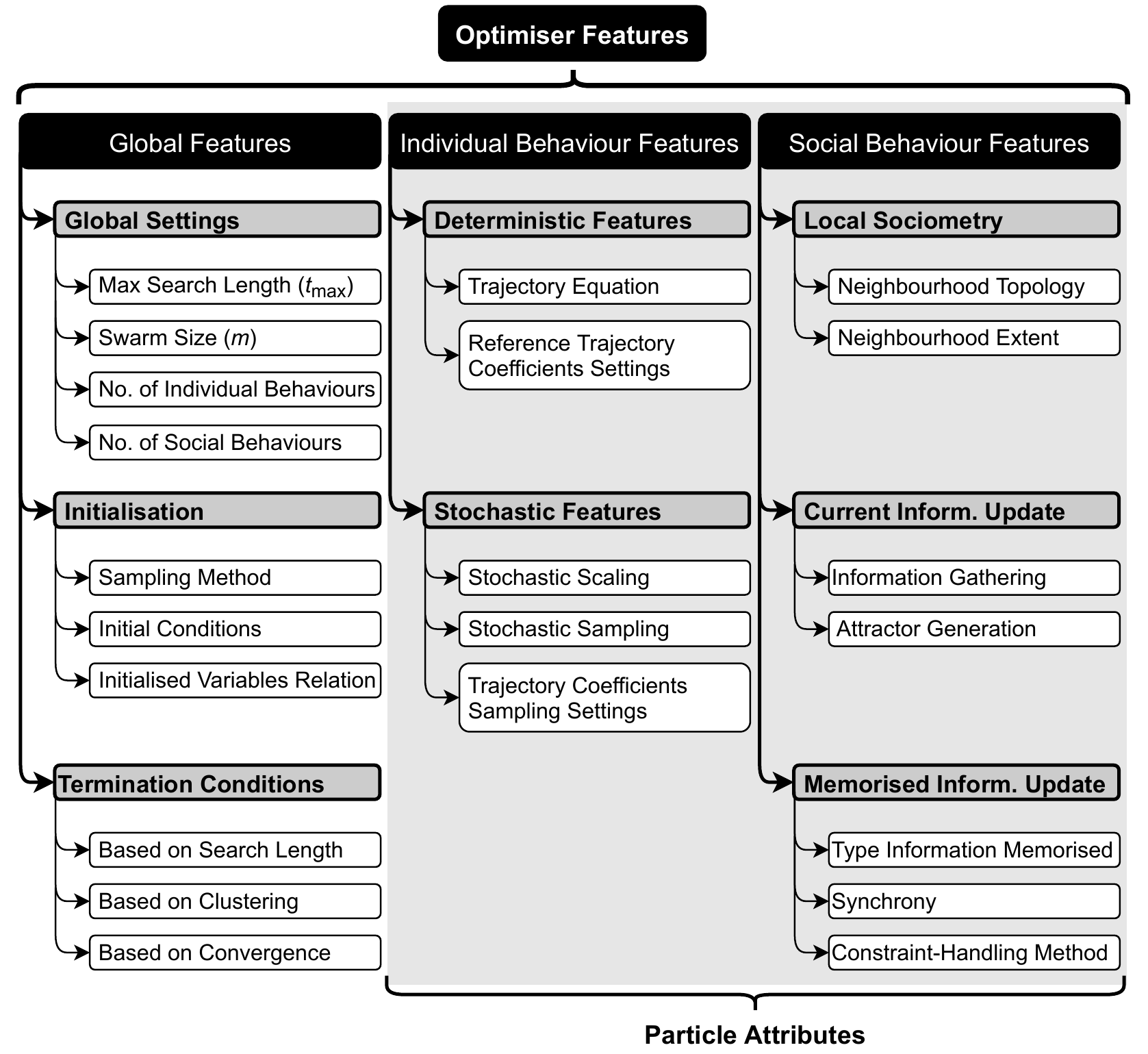}
		\caption{High-level description of the proposed Reformulated PSO (RePSO).}
	\end{center}
	\label{fig:RePSO}
\end{figure*}

The proposed \textit{Reformulated Particle Swarm Optimisation} (RePSO) method is structured in three sets of features: 1)~\textit{Global Features} (GFs), 2)~\textit{Individual Behaviour Features} (IBFs), and 3)~\textit{Social Behaviour Features} (SBFs). Fig.~\ref{fig:RePSO} shows a high-level description of RePSO, where IBFs and SBFs are both viewed as individual attributes of a particle (\textit{Particle Attributes}).


\section{Global Features}
\label{sec:1.GLobFeat}

Despite being a swarm-intelligent method, some characteristics must still be defined at the swarm level. We define here three main subsets of global features: 1)~\textit{Global Settings}, 2)~\textit{Initialisation}, and 3)~\textit{Termination Conditions}. The first one consists of scalar settings like maximum search length ($t_\textit{max}$) and swarm size ($m$), whereas the other two involve methods. Whilst the \textit{Neighbourhood Topology} and the \textit{Constraint-Handling Technique} are traditionally viewed as global features, it is porposed here that they be defined at the individual level.

\subsection{Initialisation}
\label{sec:2.Initialisation}

It is important to differentiate two aspects of the initialisation in PSO:

\begin{enumerate}
	\item The \textit{sampling method} to be used to place $m$ points over the search-space.
	\item What \textit{variables} associated with each particle are to be initialised.
\end{enumerate}

Note that the particle's position update in RePSO --and in \eqref{eq:CPSO}-- is a $2^\text{nd}$~order difference equation rather than the classical system of two $1^\text{st}$~order difference equations (position and velocity). Therefore, the \textit{variables} which may be involved in the initialisation are the initial, the previous, and the memorised positions ($\mathbf{x}^{(1)}$, $\mathbf{x}^{(0)}$, $\mathbf{xm}^{(1)}$) instead of two positions and one velocity ($\mathbf{x}^{(1)}$, $\mathbf{xm}^{(1)}$, $\mathbf{v}^{(1)}$).

\subsubsection{Sampling Method.}

Originally, initialisation was purely random from uniform distributions:
$x_{ij}^{(1)} = x_{\text{min }ij} + U_{(0,1)} \left(x_{\text{max }ij} - x_{\text{min }ij}\right)$. \textit{Random Sampling} is easy to implement but does not usually result in good coverage of the search-space. More advanced sampling methods may be used, such as \textit{Latin Hypercube Sampling}, \textit{Orthogonal Sampling} or a range of different \textit{Tesselations}.

\subsubsection{Initial Conditions.}

This aspect of the PSO method seldom receives any attention from researchers or practitioners. Four types are proposed here:

\begin{enumerate}
	\item \textit{Stagnation}: $\mathbf{x}^{(1)} = \mathbf{x}^{(0)} = \mathbf{xm}^{(1)}$\\
	This requires the \textit{sampling} of each particle's position at the initial time-step ($\mathbf{x}^{(1)}$). Stagnation implies that the previous position $\mathbf{x}^{(0)} = \mathbf{x}^{(1)}$, and that the particle has converged to its attractor: $\mathbf{xm}^{(1)}= \mathbf{x}^{(1)}$. Thus, movement starts purely due to cooperation (no inertia, no individual attractor).\medskip
	
	\item \textit{Two Positions}: $\mathbf{x}^{(1)} \neq \mathbf{x}^{(0)}$ and either $\mathbf{xm}^{(1)} = \mathbf{x}^{(1)}$ or $\mathbf{xm}^{(1)} = \mathbf{x}^{(0)}$ \\
	Two positions per particle are \textit{sampled} and compared, with the better one becoming $\mathbf{x}^{(1)}$, the other becoming $\mathbf{x}^{(0)}$, and $\mathbf{xm}^{(1)} = \mathbf{x}^{(1)}$. Thus, movement starts both due to cooperation and to inertia (no individual attractor).\medskip
	
	\item \textit{One Position and One Memory}: $\mathbf{x}^{(1)} = \mathbf{x}^{(0)} \neq \mathbf{xm}^{(1)}$ \\
	Two positions per particle are \textit{sampled} and compared, with the better one becoming $\mathbf{xm}^{(1)}$ and the other $\mathbf{x}^{(1)} = \mathbf{x}^{(0)}$. Movement starts due to both cooperation and acceleration towards its individual attractor (no inertia).\medskip
	
	\item \textit{Two Positions and One Memory} $\mathbf{x}^{(1)} \neq \mathbf{x}^{(0)} \neq \mathbf{xm}^{(1)} \neq \mathbf{x}^{(1)}$\\
	Three positions per particle are \textit{sampled} and compared, with the best one becoming $\mathbf{xm}^{(1)}$. Thus, movement starts both due to all three sources: cooperation, inertia, and acceleration towards its individual attractor.
\end{enumerate}

\subsubsection{Initialised Variables Relation.}

For all \textit{initial conditions} other than \textit{stagnation}, more than one position is to be \textit{sampled} per particle. The question is then whether these should be somehow related. Three alternatives are proposed here:

\begin{enumerate}
	\item \textit{Perturbation}: $\mathbf{x}^{(0)}$ is generated from controlled perturbations on $\mathbf{x}^{(1)}$. If applicable, $\mathbf{xm}^{(1)}$ is also generated from perturbations on $\mathbf{x}^{(1)}$.\medskip
	
	\item \textit{Independent}: Each population of positions is sampled independently.\medskip
	
	\item \textit{Simultaneous}: All populations of positions are sampled at once. For instance, if using the Latin Hypercube Sampling, there would be one single sampling with as many points as twice or three times the swarm size, as applicable.
\end{enumerate}

\subsection{Termination Conditions}
\label{sec:2.TCs}

The population-based nature of the method enables termination conditions different from the classical ones in numerical optimisation: 1)~maximum number of iterations, and 2)~convergence. Three types of conditions are identified here: 1)~\textit{based on search length} (or maximum number of iterations), 2)~\textit{based on clustering measures} (diversity loss), and 3)~\textit{based on measures of convergence}.

\section{Individual Behaviour Features}
\label{sec:1.IBFeat}

These are the features of the algorithm which control the individual behaviour of a particle. Each particle has its own set of IBFs, which are viewed as particle attributes. The individual behaviour of a particle is materialised by its trajectory as it is pulled by its attractor. This is governed by a second order difference equation and the setting of its coefficients. The IBFs are grouped here in two main families, namely \textit{Deterministic Features} and \textit{Stochastic Features}.

\subsection{Deterministic Features}
\label{sec:2.DetFeat}

Instead of viewing PSO as a \textit{guided random search method}, it is viewed as a \textit{randomly-weighted deterministic search method}. Thus, its desired deterministic behaviour is defined, adding only as much stochasticity as deemed beneficial.

By formulating the position update as in \eqref{eq:MyPSO}, it is clear that any given particle at any given time is pulled by a single attractor which results from a randomly weighted average of the components of the individual and social attractors. Thus, the \textit{Trajectory Difference Equation} in \eqref{eq:CPSO} may be expressed as in \eqref{eq:MyPSO}.

\begin{equation}
	\label{eq:SingleTerm}
	iw_{ij}^{(t)} U_{(0,1)} \left(xb_{ij}^{(t)}-x_{ij}^{(t)}\right) + sw_{ij}^{(t)} U_{(0,1)}\left(xb_{kj}^{(t)}-x_{ij}^{(t)}\right) = \phi_{ij}^{(t)} \left(p_{ij}^{(t)}-x_{ij}^{(t)}\right)
\end{equation}

\begin{equation} \label{eq:phi}
	\phi_{ij}^{(t)} = \iota_{ij}^{(t)} + \sigma_{ij}^{(t)} = iw_{ij}^{(t)} ~ U_{(0,1)} + sw_{ij}^{(t)} ~ U_{(0,1)}
\end{equation}

\begin{equation}
	\label{eq:attractor}
	p_{ij}^{(t)} = \frac{\iota_{ij}^{(t)} ~ xb_{ij}^{(t)} + \sigma_{ij}^{(t)} ~ xb_{kj}^{(t)}}{\phi_{ij}^{(t)}}
\end{equation}

\begin{equation}
	\label{eq:MyPSO}
	x_{ij}^{(t+1)} = x_{ij}^{(t)} + \omega_{ij}^{(t)} \left(x_{ij}^{(t)} - x_{ij}^{(t-1)}\right) + \phi_{ij}^{(t)} \left(p_{ij}^{(t)}-x_{ij}^{(t)}\right)
\end{equation}

\subsubsection{Trajectory Equation.}

Since we are dealing with a single particle, sub-index $i$ is dropped. For simplicity, let us assume $(\omega,\phi)$ constant in every dimension and $\forall t$, dropping sub-index $j$ and super-index $(t)$. If stochasticity is removed, the deterministic coefficients $(\hat{\omega}, \hat{\phi})$ are referred to as \textit{Reference Trajectory Coefficients}.
\\
\\
\underline{\textit{CPSO Recurrence Formulation}}. The \textit{CPSO Recurrence Formulation} is as in \eqref{eq:MyDetPSO}, which is the deterministic version of \eqref{eq:MyPSO}. The generation of the overall attractor $\textbf{p}_i^{(t)}$ is now decoupled, comprising a \textit{Social Behaviour Feature} (SBF).

\begin{equation}
	\label{eq:MyDetPSO}
	\boxed{
	x_{ij}^{(t+1)} = x_{ij}^{(t)} + \hat{\omega} \left(x_{ij}^{(t)} - x_{ij}^{(t-1)}\right) + \hat{\phi} \left(p_{ij}^{(t)}-x_{ij}^{(t)}\right)
	}
\end{equation}
\\
\\
\underline{\textit{CPSO Closed-Form Formulation}}. This is obtained by solving the difference equation in \eqref{eq:MyDetPSO}. The roots of the characteristic polynomial are as in \eqref{eq:Char Polyn Roots} and \eqref{eq:Gamma}. The solution is per dimension and per particle (therefore dropping indices $i$ and $j$), and the attractor \textbf{p} is stationary.

\begin{equation} \label{eq:Char Polyn Roots}
	r_1 = \frac{1+\hat{\omega}-\hat{\phi}}{2} + \frac{\gamma}{2} \quad;\quad r_2 = \frac{1+\hat{\omega}-\hat{\phi}}{2} - \frac{\gamma}{2}
\end{equation}

\begin{equation} \label{eq:Gamma}
	\gamma = \sqrt{\hat{\phi}^2 - \left(2  \hat{\omega} + 2\right)  \hat{\phi} + \left(\hat{\omega} - 1\right)^2}
\end{equation}
\medskip
\\
\textit{Case 1} ($\gamma\textsuperscript{2} > 0$). The two roots of the characteristic polynomial are real-valued and different ($r_1 \neq r_2$). Therefore the closed-form for \textit{Case 1} is as in \eqref{eq:Position 1}.

\begin{equation} \label{eq:Position 1}
	\boxed{x^{(t)} = p + \frac{r_2  \left(p - x^{(0)}\right) - \left(p - x^{(1)}\right)}{\gamma}  r_1^t + \frac{-r_1  \left(p - x^{(0)}\right) + \left(p - x^{(1)}\right)}{\gamma} r_2^t}
\end{equation}
\medskip
\\
\textit{Case 2} ($\gamma\textsuperscript{2} = 0$). The two roots of the characteristic polynomial are the same ($r_1 = r_2$), as shown in \eqref{eq:Same Roots}. Therefore the closed-form for \textit{Case 2} is as in \eqref{eq:Position 2}.

\begin{equation} \label{eq:Same Roots}
	r = r_1 = r_2 = \frac{1+\hat{\omega}-\hat{\phi}}{2}
\end{equation}

\begin{equation} \label{eq:Position 2}
	\boxed{
	\begin{split}
		x^{(t)} &= p + \\
		 &\left[-\left(p - x^{(0)}\right) + \left(\left(p - x^{(0)}\right) - \frac{2 \left(p - x^{(1)}\right)}{1+\hat{\omega}-\hat{\phi}}\right)  t \right] \left(\frac{1+\hat{\omega}-\hat{\phi}}{2} \right)^t
	\end{split}
	 }
\end{equation}
\medskip
\\
\textit{Case 3} ($\gamma\textsuperscript{2} < 0$). The two roots are complex conjugates.

\begin{equation} \label{eq:Char Polyn Roots Comp}
	r_1 = \frac{1+\hat{\omega}-\hat{\phi}}{2} + \left(\frac{\gamma'}{2}\right) i \quad;\quad r_2 = \frac{1+\hat{\omega}-\hat{\phi}}{2} - \left(\frac{\gamma'}{2}\right) i
\end{equation}

\begin{equation} \label{eq:Gamma'}
	\gamma' = \sqrt{-\gamma^2} = \sqrt{-\hat{\phi}^2 + \left(2 \hat{\omega} + 2\right) \hat{\phi} - \left(\hat{\omega} - 1\right)^2}
\end{equation}

Using polar coordinates ($\rho,\theta$), the closed-form for \textit{Case 3} is as in \eqref{eq:Position 3}.

\begin{equation} \label{eq:Rho Theta}
	\rho = \sqrt{\hat{\omega}} \quad;\quad 	\theta = \text{acos}\left(\frac{1+\hat{\omega}-\hat{\phi}}{2 \sqrt{\hat{\omega}}}\right)
\end{equation}

\begin{equation} \label{eq:Sin Cos}
	\cos{\left(\theta\right)} = \frac{1}{\sqrt{\hat{\omega}}} \left(\frac{1+\hat{\omega}-\hat{\phi}}{2}\right) \quad;\quad \sin{\left(\theta\right)} = \frac{1}{\sqrt{\hat{\omega}}} \left(\frac{\gamma'}{2}\right)
\end{equation}

\begin{equation} \label{eq:Position 3}
	\boxed{
		\begin{split}
		x^{(t)} &= p - \sqrt{\hat{\omega}}^t \left(p - x^{(0)}\right) \cos{\left(\theta t\right)} +\\
		&\sqrt{\hat{\omega}}^t \left(\frac{\left(1+\hat{\omega}-\hat{\phi}\right) \left(p - x^{(0)}\right) - 2 \left(p - x^{(1)}\right)}{\gamma'}\right) \sin{\left(\theta t\right)}
		\end{split}
	}
\end{equation}

Thus, the chosen trajectory equation in RePSO may be given by the \textit{Recurrence Formulation} in \eqref{eq:MyDetPSO} or by the \textit{Closed-Form Formulations} in \eqref{eq:Position 1}, \eqref{eq:Position 2} and \eqref{eq:Position 3}. Other recurrence formulations as well as some considerantions to be taken into account for the closed-form formulation are beyond the scope of this paper.

\subsubsection{Reference Trajectory Coefficients Settings.}

\begin{figure}[ht!]
	\begin{subfigmatrix}{2} 
		\subfigure{\includegraphics[width=0.48\textwidth]{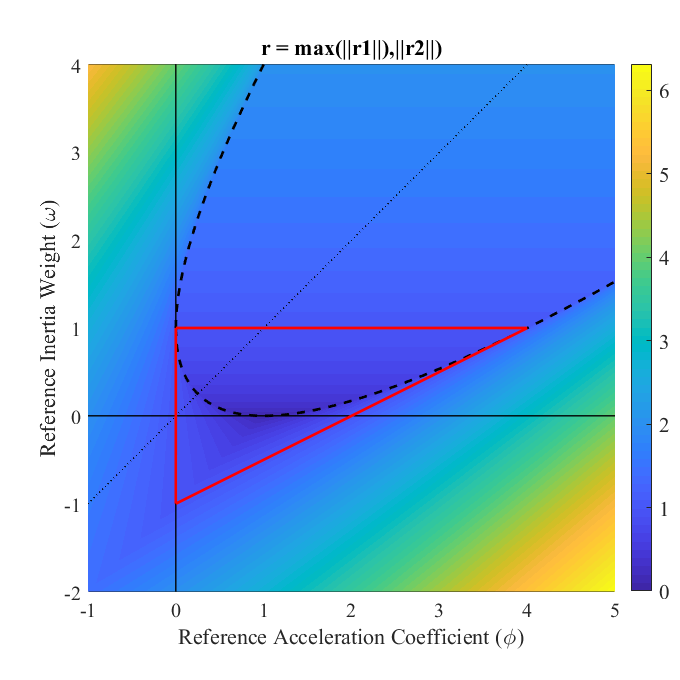}}
		\subfigure{\includegraphics[width=0.48\textwidth]{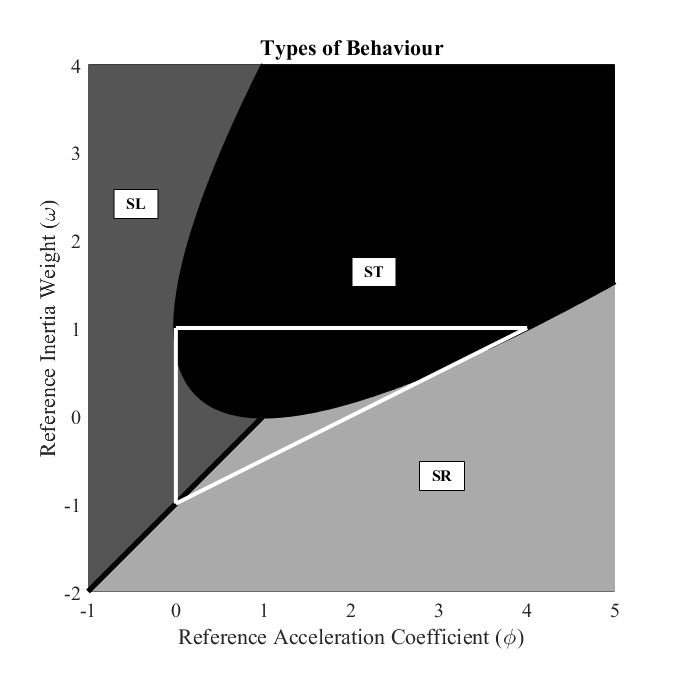}}
	\end{subfigmatrix}
	\caption{On the left, magnitude of the dominant root. Settings inside red triangle ensure convergence ($r < 1$). On the right, \textit{Sectors} for three \textit{Types of Behaviour} in CPSO: black region is \textit{Oscillatory}, dark grey region is \textit{Monotonic}, and light grey region is \textit{Zigzagging}.}
	\label{fig:TypeBehaviour}
\end{figure}

An analysis of the trajectory closed-forms shows that the magnitude of the dominant root $r = \text{max}\left(\left\|r_1\right\|,\left\|r_2\right\|\right)$ controls convergence. Fastest convergence occurs for $(\hat{\phi},\hat{\omega})$ = $(1,0)$, where $r = 0$ (see Fig.~\ref{fig:TypeBehaviour}~(a)). The resulting convergence conditions are shown in \eqref{eq:Convergence Conditions}, which define the area inside the convergence triangle ($r < 1$) shown in Fig.~\ref{fig:TypeBehaviour}.

\begin{equation} \label{eq:Convergence Conditions}
	\boxed{\begin{array}{l}
			\hat{\omega} < 1 \\
			\hat{\phi} > 0 \\
			\displaystyle \hat{\omega} > \frac{\hat{\phi}}{2} - 1
	\end{array}}
\end{equation}

Whilst the magnitude of the dominant root controls the speed of convergence, the sign of the dominant root (if any) controls the \textit{Type of Behaviour}:

\begin{enumerate}
	\item \textit{Oscillatory}: Roots are complex conjugates (no dominant root).
	\item \textit{Monotonic}: Dominant root is real-valued and positive.
	\item \textit{Zigzagging}: Dominant root is real-valued and negative.
\end{enumerate}

These \textit{Types of Behaviour} are bounded within specific \textit{Sectors} in the $(\hat{\omega},\hat{\phi})$ plane, each associated with one edge of triangular isolines (same $r$). These three \textit{Sectors} are shown in Fig.~\ref{fig:TypeBehaviour}~(b), where the white triangle separates the \textit{Convergence} (inside) and \textit{Divergence} regions. The settings of ($\hat{\omega}, \hat{\phi}$) can be chosen so as to achieve the desired behaviour and convergence speed. For example:
\begin{enumerate}
	\item Choose \textit{Type of Behaviour}: e.g. \textit{Oscillatory}.
	\item Set \textit{Convergence Speed}: $\sqrt{\hat{\omega}} \in [0,1]$, with fastest convergence for $\sqrt{\hat{\omega}} = 0$.
	\item Set \textit{Reference Acceleration Coefficient}: $\hat{\phi} \in \left(\left(\sqrt{\hat{\omega}}-1\right)^2,\left(\sqrt{\hat{\omega}}+1\right)^2\right)$.

\end{enumerate}

\subsection{Stochastic Features}
\label{sec:2.StochFeat}

The random weights in \eqref{eq:CPSO} affect the trajectory of a particle towards the overall attractor (\textbf{p}) whilst also affecting the generation of this attractor as a stochastic convex combination of the individual ($\mathbf{xb}_i$) and the social ($\mathbf{xb}_k$) attractors. Different from classical PSO formulations, these two features are decoupled here. The \textit{Stochastic Features} within the IBFs are concerned only with the former.

\subsubsection{Stochastic Scaling.}

This refers to whether the stochastic variables in \eqref{eq:MyPSO} are sampled once per particle position update (\textit{vector scaling}) or resampled anew per dimension as well (\textit{component scaling}). The latter is sometimes used by mistake.

\subsubsection{Stochastic Sampling.}

This refers to the probability distributions chosen for the control coefficients $(\omega,\phi)$. In classical formulations, $\omega = \hat{\omega}$ (deterministic) whereas the probability distribution of $\phi$ is not directly chosen but resulting from the sum of two stochastic terms sampled from uniform distributions: $\phi = \iota + \sigma$ as in \eqref{eq:phi}. If they are sampled from the same interval (i.e. $iw = sw$), the resulting probability distribution of $\phi$ is triangular. Otherwise, it is trapezoidal. In the proposed RePSO, the user is allowed to choose any distribution for $(\omega,\phi)$.

\subsubsection{Trajectory Coefficients Sampling Settings.}

Once the distributions have been chosen, the parameters defining them must be set. For example, $\left(\phi_\text{min},\phi_\text{max}\right)$ for a uniform distribution, or the standard deviation for a normal distribution.

\section{Social Behaviour Features}
\label{sec:1.SBFeat}

These are the features of the algorithm which control the social behaviour of a particle. Despite being SBFs, they are defined as \textit{Particle Attributes} in RePSO. A particle's social behaviour is governed by its access to other particles' memories (\textit{Local Sociometry}) and by how it handles this information (social influence).

\subsection{Local Sociometry}
\label{sec:2.LocalSoc}

In classical formulations, the sociometry is a global feature. It can be defined as a regular graph, or irregulary by defining one connection at a time. In the latter case, the structure cannot be automatically generated nor is it scalable. In RePSO, automation and scalability are paramount. Therefore, a \textit{Local Sociometry} is defined for each particle, with the \textit{Global Sociometry} resulting from their assembly. This has the advantange that sociometry is a particle attribute, facilitating object-oriented implementation. Also that different social behaviours can be exhibited by different particles, and that irregular global sociometries are possible without renouncing automation or scalability.

\begin{figure}[h!]
	\begin{subfigmatrix}{2} 
		\subfigure[Sociometry]{\includegraphics[width=0.23\textwidth]{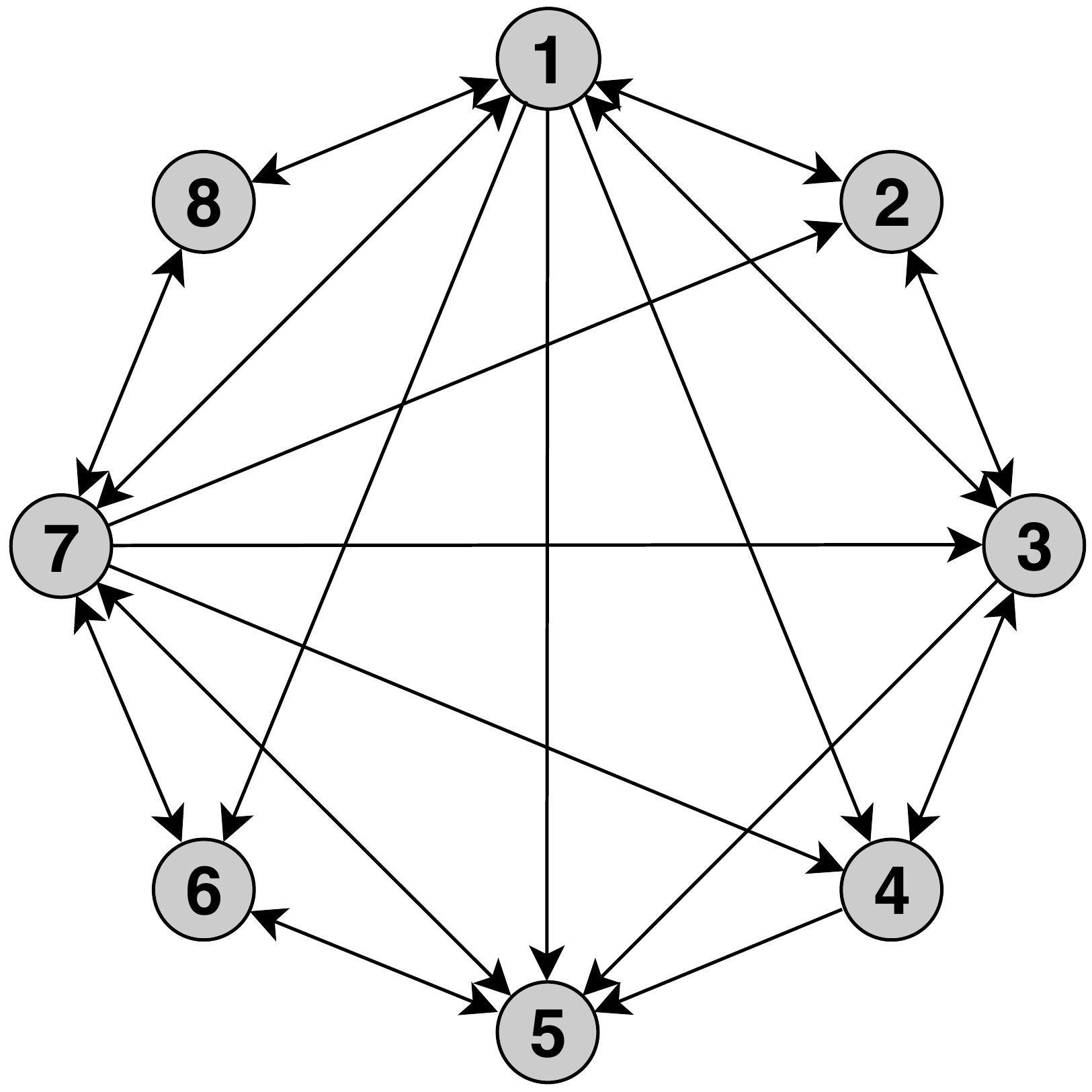}}
		\subfigure[Connectivity Matrix]{\includegraphics[width=0.44\textwidth]{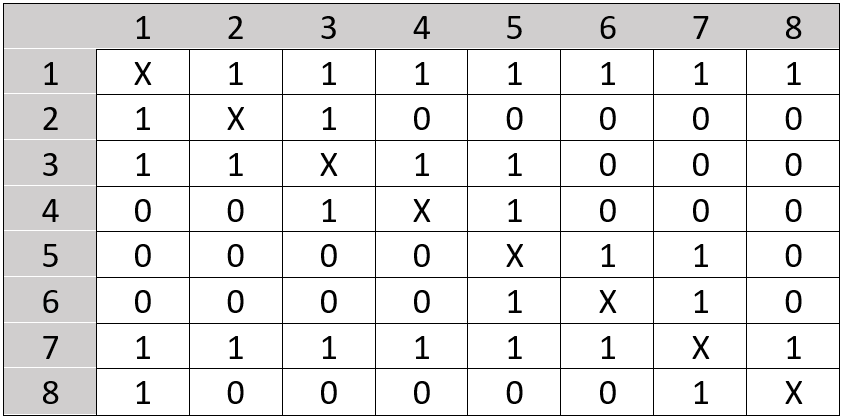}}
	\end{subfigmatrix}
	\caption{Unstructured sociometry emerging from local sociometries.}
	\label{fig:MoreTopologies}
\end{figure}

The Local Sociometry in RePSO is generated by defining the \textit{Neighbourhood Topology} and potentially the \textit{Neighbourhood Extent}. Examples of the former are the \textit{Global}, \textit{Ring}, \textit{Forward} and \textit{Wheel} topologies. The \textit{Neighbourhood Topology} defines a methodology to generate unidirectional connections from the particle informed to its informers. The \textit{Neighbourhood Extent} defines the neighbourhood size, if applicable (number of neighbours, distance of influence). An example of an unstructured global neighbourhood is shown in Fig.~\ref{fig:MoreTopologies}, where the Local Sociometry of particle~1 is the \textit{Global} topology whilst that of particle~2 is the \textit{Ring} topology. Other aspects may be considered, such as whether a particle's memory is part of its neighbourhood (X in the connectivity matrix in Fig.~\ref{fig:MoreTopologies}).

\subsection{Current Information Update}
\label{sec:2.CurrentInfo}

Any particle holds two types of information: 1) \textit{current}, and 2) \textit{memorised}. The update of the former takes place by gathering information (e.g. memorised attractors), generating an overall attractor using the information gathered, and then applying the trajectory equation to update the current information. A particle may access the information currently held, the one memorised, or both from its neighbours (\textit{Information Gathering}). This is an extension to classical formulations, where a particle can only access their memorised information.

\subsection{Memorised Information Update}
\label{sec:2.MemInfo}

This controls the update of a particle’s memory when it accesses new information. This is performed directly rather than through a trajectory equation. The question is what \textit{Type of Information} is accessible to a particle’s memory.

Another feature affecting this update is the \textit{Synchrony}, which defines whether a particle's memory is updated immediately after its currently held information is updated (asynchronous) or only after the currently held information of every particle is updated (synchronous). Typically, the update is synchronous.

RePSO also proposes to include the CHTs here. Thus, different particles may have different CHTs, and therefore may value a given location differently.

\section{Conclusions}
\label{sec:1.Conclusions}

A general framework has been proposed aiming to encompass many variants of the PSO algorithm under one umbrella so that different versions may be posed as settings within the proposed unified framework. In addition, some extensions to the classical PSO method have been made such as the decoupling of the stochasticity that affects both the acceleration coefficient ($\phi$) and the generation of the overall attractor, an extended treatment of the swarm initialisation, the particle trajectory closed forms, the identification of three types of deterministic behaviour to inform the setting of the control coefficients, and the global sociometry resulting from assembling local sociometries defined as particle attributes. Due to space constraints, most of these features are discussed only superficially.

%
%
\bibliographystyle{splncs04}
\bibliography{BibliographyRePSO}
%
%




%
\end{document}